\documentclass[conference]{IEEEtran}
\IEEEoverridecommandlockouts
\usepackage{tcolorbox}
\usepackage{multirow}
\usepackage{mathtools}
\usepackage{enumitem}
\usepackage{cite}
\usepackage{amsmath,amssymb,amsfonts}
\usepackage{algorithmic}
\usepackage{graphicx}
\usepackage{textcomp}
\usepackage{xcolor}
\usepackage{hyperref}
\usepackage{caption}
\DeclarePairedDelimiter\ceil{\lceil}{\rceil}
\def\BibTeX{{\rm B\kern-.05em{\sc i\kern-.025em b}\kern-.08em
    T\kern-.1667em\lower.7ex\hbox{E}\kern-.125emX}}
\begin{document}

\title{Cross-lingual COVID-19 Fake News Detection}

\author{\IEEEauthorblockN{Jiangshu Du\IEEEauthorrefmark{1},
Yingtong Dou\IEEEauthorrefmark{1},
Congying Xia\IEEEauthorrefmark{1}, 
Limeng Cui\IEEEauthorrefmark{2},
Jing Ma\IEEEauthorrefmark{3} and Philip S. Yu\IEEEauthorrefmark{1}}
\IEEEauthorblockA{\IEEEauthorrefmark{1}Department of Computer Science, University of Illinois at Chicago, USA}
\IEEEauthorblockA{\IEEEauthorrefmark{2}Department of Computer Science and Engineering, Pennsylvania State University, USA}
\IEEEauthorblockA{\IEEEauthorrefmark{3}Department of Computer Science, Hong Kong Baptist University, HongKong\\ \{jdu25, ydou5, cxia8, psyu\}@uic.edu, lzc334@psu.edu, majing@hkbu.edu.hk}}

\maketitle

\begin{abstract}
The COVID-19 pandemic poses a great threat to global public health.
Meanwhile, there is massive misinformation associated with the pandemic which advocates unfounded or unscientific claims.
Even major social media and news outlets have made an extra effort in debunking COVID-19 misinformation,
most of the fact-checking information is in English, whereas some unmoderated COVID-19 misinformation is still circulating in other languages, threatening the health of less-informed people in immigrant communities and developing countries.
In this paper, we make the first attempt to detect COVID-19 misinformation in a low-resource language (Chinese) only using the fact-checked news in a high-resource language (English). 
We start by curating a Chinese real\&fake news dataset according to existing fact-checking information.
Then, we propose a deep learning framework named $\mathsf{CrossFake}$ to jointly encode the cross-lingual news body texts and capture the news content as much as possible.
Empirical results on our dataset demonstrate the effectiveness of $\mathsf{CrossFake}$ under the cross-lingual setting and 
it also outperforms several monolingual and cross-lingual fake news detectors.
The dataset is available at \url{https://github.com/YingtongDou/CrossFake}. 
\end{abstract}


\section{Introduction}

The booming of social media fosters various kinds of misinformation where fake news is a typical form that represents news articles that are intentionally and verifiably false~\cite{zhou2020survey}.
Amid the unprecedented COVID-19 global pandemic, the mysterious cause of the coronavirus and its severe infectiousness have incited more fake news.
For instance, many news articles falsely claim that drinking or injecting bleach could kill the COVID-19 virus, which has already made detrimental hurt to less-informed individuals\footnote{\url{https://nyti.ms/3j8E7OJ}}.
The WHO has officially declared the wide-spreading COVID-19 misinformation as a ``\textit{infodemic}'' and called for mitigating it~\cite{who2020infodemic}.

Fact-checking by domain experts is a widely used approach to debunk COVID-19 misinformation.
Besides professional fact-checking organizations like Politifact\footnote{\url{https://www.politifact.com/coronavirus/}}, the infodemic has prompted many social media, news outlets, and even governments to launch exclusive tools to debunk misinformation~\cite{siwakoti2021covid}.
However, recent reports\footnote{\url{https://nyti.ms/3gqujOl}, \url{https://bit.ly/38BYZI7}} notice that the COVID-19 misinformation has imposed threats to non-English speakers.
Since those people do not consume English media, the vetted information in English is hardly accessed by them.
Meanwhile, the lack of fact-checking or content moderation in some non-English media exacerbates the negative influence of misinformation.
A case we have discovered is an English post falsely claiming that U.S. hospitals are preparing for 96 million coronavirus infections at the early phase of the pandemic\footnote{\url{https://bit.ly/3keGK0L}}, many news articles claiming the same thing are still existing on various Chinese social media platforms\footnote{\url{https://archive.vn/SVGuK}, \url{https://archive.vn/FOidI}} after one year the source being debunked.
Thus, it is imperative to develop an effective fake news detection model for low-resource languages.

Among previous fake news detection works~\cite{zhou2020survey}, only a few investigated fake news under the cross-lingual or multi-lingual setting.
Some papers adopted pre-trained multi-lingual encoders to encode the news in different languages~\cite{dementieva2020fake, schwarz2020emet}; while a recent work utilized language-independent features to handle the multi-lingual setting~\cite{vogel2020detecting}, and another work applied transfer learning to map the monolingual word embeddings from different languages into the same space~\cite{haider2020detecting}.
Besides the above works, several cross-lingual learning approaches have been applied to similar domains like hate speech~\cite{stappen2020cross} and abusive language detection~\cite{glavavs2020xhate}.

Regarding the cross-lingual COVID-19 infodemic, the annotated news articles in low resource languages are scarce, and the news develops quickly across different languages with many new terms.
Therefore, it is infeasible to train a monolingual model based on a low-resource language with few annotations.
Moreover, the lack of news social engagement information in some low-resource languages impedes the application of social context-based fake news detectors~\cite{nguyen2020fang}.

To cope with the above challenges, we attempt to train a cross-lingual fake news detector trained solely based on a high resource language (English) COVID-19 news corpus and used to predict news credibility in a low resource language (Chinese).
We start by curating a COVID-19 news dataset in Chinese based on existing fact-checking information.
Then, we propose an end-to-end fake news detection framework named $\mathsf{CrossFake}$ based on pre-trained language models.
To deal with the long news body text, it is sliced into sub-text groups before being fed into language encoders.
Experimental results verify the effectiveness of proposed $\mathsf{CrossFake}$ comparing to monolingual and cross-lingual baselines.
We also discuss the advantage and drawbacks of the proposed model using several real-world cases.
We summarize our contributions as follows:

\begin{itemize}
\item We collect and annotate a fine-grained cross-lingual COVID-19 fake news dataset.

\item We propose an end-to-end cross-lingual fake news detector tailored to the news text properties.

\item We empirically show the advantage and limitation of $\mathsf{CrossFake}$ comparing to mono/cross-lingual baselines.

\end{itemize}

\section{Data Collection}
\label{sec02-data}

Recent works~\cite{zhou2020recovery, cui2020coaid, li2020mm, shahi2020fakecovid} have released several annotated English COVID-19 news datasets according to the fact-checking information from recognized fact-checking institutions.
In contrast, there is seldom fact-checked Chinese COVID-19 news corpus.
Although~\cite{yang2020checked} has curated a fact-checked Chinese COVID-19 misinformation dataset, it only contains short-text social media posts without complete news body text.
We also notice some COVID-19 fact-checking websites operated by Chinese social media or government\footnote{\url{https://www.piyao.org.cn/2020yqpy/}, \url{https://vp.fact.qq.com/home}}.
However, their fact-checking articles only analyze and verify specific claims related to COVID-19 without referring to their corresponding source news articles.

To obtain high-quality news data, we resort to annotate Chinese news manually.
Given thousands of annotated English COVID-19 news in existing datasets~\cite{zhou2020recovery, cui2020coaid},
we find their corresponding Chinese news as the annotated Chinese news and our data collection protocol is shown below:

\small
\begin{tcolorbox}[colframe=gray,colback=white,boxrule=1pt,arc=3pt,boxsep=2pt,left=-1pt,right=3pt,top=3pt,bottom=3pt]
    
\textbf{Step 1.} Select a piece of English news in existing datasets.

\vspace{1mm}
\textbf{Step 2.} Search the translated English title under three major Chinese news search engines\footnote{\url{https://sogou.com/}, \url{https://toutiao.com/}, \url{http://baidu.com/}}.

\vspace{1mm}
\textbf{Step 3.} Check if there is a news title on the first page of search results that has a similar meaning as the original English news. If yes, go to Step 4; else, go to Step 1.

\vspace{1mm}
\textbf{Step 4.} Check if the content of the selected Chinese news and original English news express similar opinions towards the same event/claim. If yes, go to Step 5; else, go to Step 1.

\vspace{1mm}
\textbf{Step 5.} Collect the metadata of the selected Chinese news and add them to the dataset.

\end{tcolorbox}

\normalsize
Four data annotators with both English and Chinese knowledge are trained with the above protocol before collecting the data.
Two annotators collect the fake news and the other two collect real news.
All annotated news articles are discussed together with an independent fact-checking expert to guarantee accuracy and consistency.
We have inspected more than one thousand fact-checked news spanning from January 2020 to October 2020 and have collected 86 fake and 114 real Chinese news.
Note that even our dataset contains news metadata like title, author, and timestamp, we only investigate the news body text in this paper for simplicity.
More dataset details are shown in Section~\ref{sec04:dataset}.

\section{Methodology}
\begin{figure}
    \centering
    \includegraphics[width=0.5\textwidth]{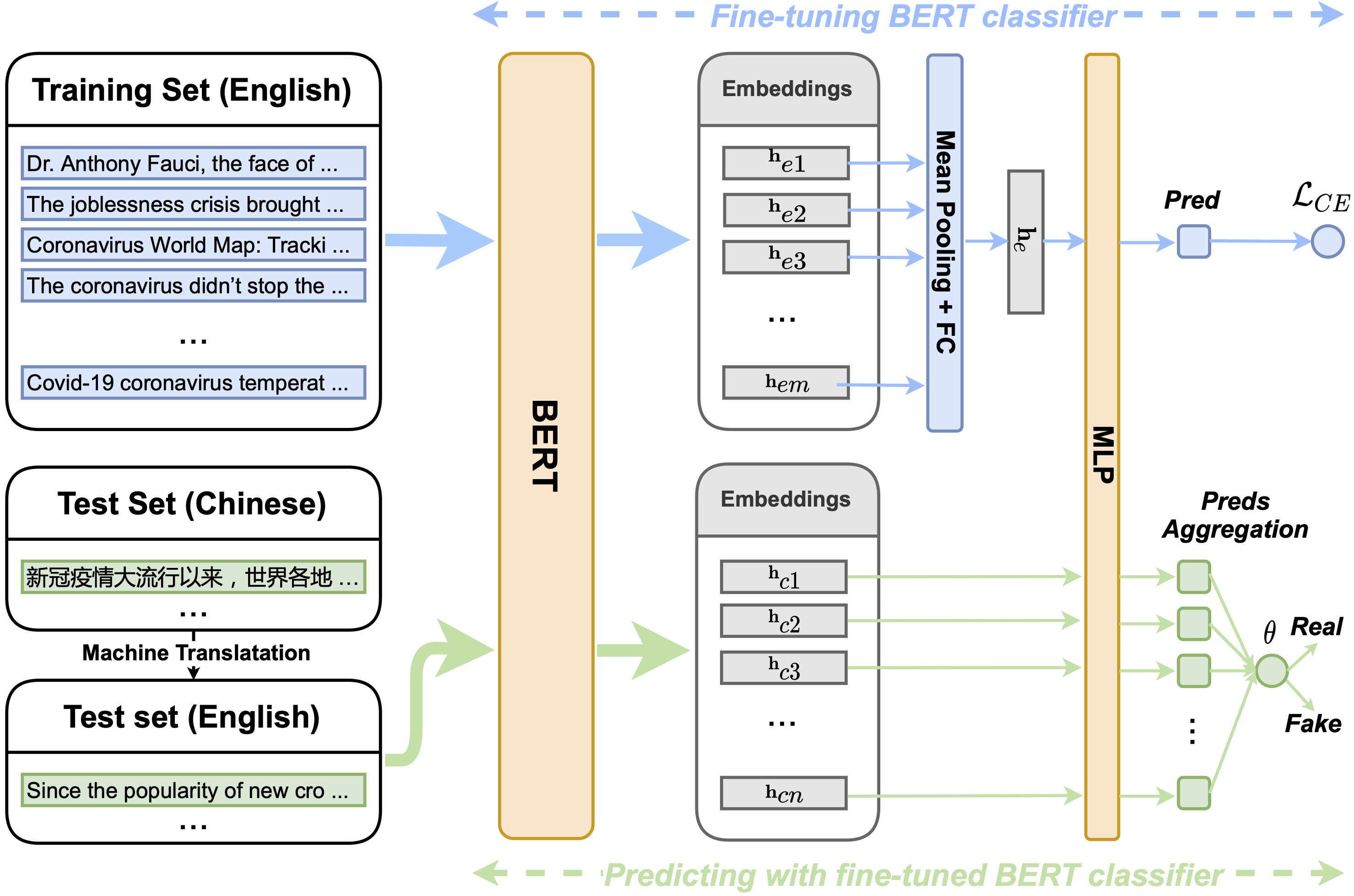}
    \caption{\small The workflow of the proposed $\mathsf{CrossFake}$ fake news detector.
    We train a neural classifier based on the aggregated BERT embeddings of a fact-checked English news sub-text (i.e., token groups)).
    To verify a Chinese news article, it is first translated into English, and the final predictions are made via aggregating all sub-text predictions.}
    \label{fig:workflow}
\end{figure}

\noindent \textbf{Problem Definition.}
This paper aims to predict the Chinese COVID-19 news truthfulness, while only a small number of annotated Chinese COVID-19 fake news is not enough to train a good supervised classifier.
Thanks to the rich annotated English COVID-19 news corpus~\cite{zhou2020recovery, shahi2020fakecovid, cui2020coaid}, we formulate our problem as a cross-lingual fake news detection task.
Specifically, we define a set of English news articles as $e \in N_{e}$ ($c \in N_{c}$ for Chinese news, resp.) and their labels as $y_{e} \in Y_{e}$ ($y_{c} \in Y_{c}$ for Chinese news, resp.).
The label for each news belongs to either 1 (fake news) or 0 (real news). 
Our objective is to train a classifier $C$ with training data $N_e$ and label $Y_e$ and maximize the test accuracy of $C$ on $N_c$ and $Y_c$.
The framework of the proposed $\mathsf{CrossFake}$ model is shown in Figure~\ref{fig:workflow}.
Next, we elucidate how to train $\mathsf{CrossFake}$ and how it verifies Chinese news credibility.

\vspace{1mm}
\noindent \textbf{Model Training.}
In the training phase. we encode the annotated English news using BERT~\cite{devlin-etal-2019-bert},
a language pre-training mode, as our base model.
Comparing to social media posts, the news body text length is usually longer.
As per our collected dataset, most of news has more than 512 tokens after tokenization, which exceeds the maximum tokens BERT can process.
To tackle this problem, we decide to break the long body text into \textit{sub-text} groups.
Specifically, given the tokenized body text $T_e$ for a piece of news $e$, we break $T_e$ into a set of $m$ token groups $TG_e = \{t_{e1}, \dots, t_{em}\}$ sequentially where $m = \ceil*{\frac{|T_e|}{500}}$.
It is to say that one sub-text size is 500.
Then, we can represent the final news embedding $\mathbf{h}_e$ as:
\begin{equation}
    \mathbf{h}_e = \mbox{FC}\left(\frac{\sum_{i=1}^{m} \mbox{BERT}(t_{ei})}{m}\right),
\end{equation}
where each token group $t_{ei} \in TG_e$ (sub-text) is encoded by BERT separately.
A mean pooling layer and a fully-connected (FC) layer are applied over all sub-texts to yield the final news embedding $\mathbf{h}_e$.
The fact-related information in news body text can be captured and retained as much as possible through the operation above.

Since our data only has fake and real news, we adopt the binary cross-entropy loss function to update the classifier $C$:
\begin{equation}\label{eq:loss}
    \mathcal{L} = \sum_{e\in \mathcal{N}_e}-\log\left(y_{e}\cdot\mbox{ReLU}(\mbox{MLP}(\mathbf{h}_{e}))\right).
\end{equation}
The loss function is optimized using SGD, and it is equivalent to fine-tuning the pre-trained BERT encoder.

\vspace{1mm}
\noindent \textbf{News Verification.}
As we mentioned in the problem definition, we need to verify the truthfulness of Chinese news based on the classifier $C$ trained in English.
As Figure~\ref{fig:workflow} illustrated, we first translate the test data (i.e., Chinese COVID-19 news) into English with Google Translator API\footnote{\url{https://pypi.org/project/google-trans-new/}} to align the input data language of $C$.
We compared baselines that encode the Chinese directly, and their performance is worse than our approach (see more in Section~\ref{sec04:analysis}).
The test data tokenizer and encoder are the same as what is used on the training data.
The sub-texts set for news $c$ is $TG_c = \{t_{c1}, \dots, t_{cn}\}$ where $n = \ceil*{\frac{|T_c|}{100}}$.
Note that we use 100 instead of 500 as the sub-text size for test data since the collected Chinese news articles are relatively shorter than English news (see Long Text\% in Table~\ref{tab:data}).
The prediction result $p_c$ for the test news $c$ is obtained via the following equation:
\begin{equation}
    p_c = \left\{\begin{array}{l} 1, if \; \frac{\sum_{j=1}^{n} \left|C(t_{cj})\right|}{n} \geq \theta, \\
    0, if \; \frac{\sum_{j=1}^{n} \left|C(t_{cj})\right|}{n} < \theta, \\
    \end{array}\right.
\end{equation}
where we aggregate the prediction results of all sub-texts for $c$ and $\theta$ is a classification threshold empirically set to 0.8.

\section{Experiment}
\label{sec04-exp}

\subsection{Dataset}
\label{sec04:dataset}
In this paper, only English news articles are used for the training purpose since we aim to leverage the fact-checking knowledge in a high resource language to verify news truthfulness in a low resource language (Chinese).
Our training dataset consists of all English COVID-19 news from three datasets: ReCOVery~\cite{zhou2020recovery}, FakeCovid~\cite{shahi2020fakecovid}, and CoAID~\cite{cui2020coaid}.
We evaluate the performance of proposed $\mathsf{CrossFake}$ on the Chinese COVID-19 news dataset collected in this paper as described in Section~\ref{sec02-data}.
Table~\ref{tab:data} shows the statistics of the training and test datasets.



\begin{table}
\centering
\small
\caption{\small Dataset statistics. Long Text\% means the percentage of news articles exceeding 512 tokens after tokenization.}
\resizebox{0.98\linewidth}{!}{%
\begin{tabular}{cccccc} 
\hline
\textbf{Dataset} & \textbf{Time}                & \textbf{Lang.} & \textbf{Long Text\%} & \textbf{Fake} & \textbf{Total}  \\ 
\hline
Training    & \multirow{2}{*}{Jan. - Oct. 2020} & ENG           & 81.87\%             & 49.23\%                       & 2840            \\
Test	&                                   & CHN           & 41.00\%             & 43.00\%                       & 200             \\
\hline
\end{tabular}}
\label{tab:data}
\end{table}

\subsection{Experimental Setup}

\noindent \textbf{Baselines}
We compare our proposed model together with several monolingual and cross-lingual fake news detectors:
\begin{itemize}[leftmargin=*]

\item \textit{Monolingual}: CSI~\cite{ruchansky2017csi}, SAFE~\cite{zhou2020safe}, and exBAKE~\cite{jwa2019exbake} are three monolingual baselines.
CSI employs an LSTM to encode the news content to detect fake news.
SAFE uses the TextCNN~\cite{zhang2015sensitivity} to encode the news textual information.
exBAKE utilizes the vanilla BERT as the English text encoder.
We train the models above on the English training data and evaluate them on the translated test data.

\item \textit{Cross-lingual}: CLEF~\cite{dementieva2020fake} and EMET~\cite{schwarz2020emet} are two cross-lingual fake news detection models.
CLEF leverages Multilingual-BERT~\cite{devlin-etal-2019-bert} to encode non-English news, and we adopt Multilingual-BERT to encode both English and Chinese data in the experiment.
EMET proposes a framework to detect misleading social media posts across different languages with the multilingual transformer~\cite{yang2019multilingual}.
In our experiment, EMET only encodes news articles since our dataset does not include other data types.

\end{itemize}

\noindent \textbf{Experimental Settings}
CSI, SAFE, exBAKE, CLEF, and $\mathsf{CrossFake}$ are implemented using PyTorch.
For EMET, we use the author's code~\cite{schwarz2020emet}.
All baselines use the batch size of 16, and $\mathsf{CrossFake}$ is trained with batch size 1.

\begin{table}
\centering
\caption{\small The fake new classification performance with standard deviation over five runs.
$\mbox{-}sub$ ($\mbox{-}avg$ resp.) represents that the final prediction is made via aggregating the prediction result of each sub-text (predicting based on the average embedding of sub-texts resp.).}
\label{tab:exp}
\begin{tabular}{l|c|ccc} 
\hline
\textbf{Model}                         & \textbf{Accuracy}  & \textbf{Precision} & \textbf{Recall} & \multicolumn{1}{c}{\textbf{F1}}  \\ 
\hline
$\text{CLEF\cite{dementieva2020fake}}$ & $43.12_{0.41}$ & $42.88_{0.43}$     & $97.38_{3.89}$  & $59.53_{1.15}$                   \\ 
\hline
$\text{EMET\cite{schwarz2020emet}}$    & $45.90_{3.29}$ & $42.15_{1.59}$     & $70.93_{18.47}$ & $51.89_{7.35}$                   \\ 
\hline
$\text{CSI\cite{ruchansky2017csi}}$    & $68.30_{1.29}$ & $61.41_{1.77}$     & $71.16_{5.01}$  & $65.81_{2.02}$                   \\ 
\hline
$\text{SAFE\cite{zhou2020safe}}$       & $71.60_{2.71}$ & $63.69_{3.80}$     & $80.70_{3.72}$  & $71.01_{1.45}$                   \\ 
\hline
$\text{exBAKE~\cite{jwa2019exbake}}$   & $64.30_{3.53}$ & $55.64_{3.99}$     & $92.09_{7.87}$  & $68.96_{0.60}$                   \\
$\text{exBAKE\mbox{-}}sub$             & $66.80_{2.91}$ & $59.73_{3.05}$     & $70.47_{10.34}$ & $64.30_{4.88}$                   \\ 
\hline\hline
$\mathsf{CrossFake}\mbox{-}avg$        & $73.60_{2.31}$ & $64.84_{2.81}$     & $85.35_{5.91}$  & $73.51_{2.18}$                   \\
$\mathsf{CrossFake}\mbox{-}sub$        & $75.00_{3.94}$ & $71.45_{5.14}$     & $70.47_{7.56}$  & $70.67_{5.12}$                   \\
\hline
\end{tabular}
\end{table}

\subsection{Result Analysis}
\label{sec04:analysis}

The average fake news classification performance with standard deviations over five random runs is reported in Table~\ref{tab:exp}, $\mbox{-}sub$ and ($\mbox{-}avg$ represent two prediction approaches (predicting-aggregating and aggregating-prediction, respectively).
From Table~\ref{tab:exp}, we have following conclusions: 

\vspace{1mm}
\noindent \textbf{1)} $\mathsf{CrossFake}$ \textbf{has the best performance.}
$\mathsf{CrossFake}\mbox{-}avg$ outperforms all baselines in accuracy and precision, $\mathsf{CrossFake}\mbox{-}sub$ obtains a further performance gain via aggregating prediction results.
The good performance of $\mathsf{CrossFake}\mbox{-}avg$ benefits from the knowledge captured from long news articles,
whereas exBAKE can only process the first 512 tokens of an article.
By leveraging the average embedding of a long news text, the original news information can be better retained.
Meanwhile, we can observe that aggregating sub-text predictions could help alleviate the bias induced by a classifier with a higher preference for fake news.
Therefore, $\text{exBAKE\mbox{-}}sub$ has better accuracy and precision than exBAKE.

\vspace{1mm}
\noindent \textbf{2)} \textbf{CNN-based model outperforms the RNN-based model.}
SAFE outperforms CSI significantly in terms of accuracy (71.60\% and 68.30\%).
It might be because sequential models like RNN and LSTM used in CSI; experience information forgetting in long sequences, which are prevalent in the news corpus.
On the contrary, SAFE adopts CNN, which could extract the local critical information related to fact-checking.

\vspace{1mm}
\noindent \textbf{3)} \textbf{Pre-trained multi-lingual models are ineffective.}
EMET and CLEF can encode Chinese news without translation.
However, both of them perform poorly on the test set.
CLEF's high false-positive rate (i.e., high recall, low precision) suggests it has an overfitting issue.
Two reasons for the poor performance of EMET and CLEF are summarized below: 

\begin{enumerate}[leftmargin=*]
\item Multilingual-BERT and the multilingual-transformer adopted by CLEF and EMET are pre-trained on the standard corpora (Wikipedia, Reddit, etc.) and released before the COVID-19 pandemic.
Therefore, they might lack the domain knowledge for the new events, and it is difficult to map the terms like ``COVID-19'' and their corresponding cross-lingual words to a similar space.
It suggests that the pre-trained language models should be kept up-to-date to handle the emerging events.

\item Similar to exBAKE, the maximum input sequence length limits the representation capability of Multilingual-BERT and multilingual-transformer while there are many long news articles in our dataset.
Besides slicing text like $\mathsf{CrossFake}$, more effective approaches are demanded to encode the long news text.
\end{enumerate}

\section{Discussion and Limitation}
\noindent \textbf{Translation Quality.}
We find that machine translation quality is a bottleneck of cross-lingual tasks, especially in emerging events.
For instance, as a novel term, the Chinese expression of ``Coronavirus'' is mistranslated as ``new crown virus'', misleading the fake news classifiers.
Moreover, we have attempted to translate all English training data to Chinese and train a Chinese fake news classifier.
The test performance is bad since the low translation quality harms the training data quality.

\vspace{1mm}
\noindent \textbf{Information Location.}
A piece of fake news may present false information in the middle or at the end of its body text.
For a fake news article discussing coconut oil can destroy Coronavirus~\footnote{\url{https://bit.ly/3sFd1lv}, \url{https://bit.ly/2WeC4Qr}}, the annotated misinformation appears after a lengthy introduction, which exceeds the maximum sequence length most language models can process.
Consequently, the fact-related information in this news will be discarded by those models.
Therefore, a model that can capture an article's complete information is crucial for automatic fact-checking.

\vspace{1mm}
\noindent \textbf{Dataset Size.}
Due to the difficulty of data collection, our test dataset size is relatively small compared to other fake news datasets.
Though the results may not be generalizable, we hope our preliminary exploration and experimental results could encourage future works in this direction.

\section{Conclusion and Future Work}
In this paper, we make the first attempt to detect COVID-19 fake news under a cross-lingual setting.
We collect and annotate a Chinese COVID-19 news dataset and propose an end-to-end fake news detector, $\mathsf{CrossFake}$, which is trained on English news and could detect most of the collected Chinese fake news after translation.
Our experimental results demonstrate the advantage of encoding more news content and the limitation of pre-trained multi-lingual encoders.
We will release the annotated data with rich metadata and encourage future research on cross-lingual fake news detection, especially in designing better models to capture the knowledge across different languages.
Moreover, the event-centric analysis based on our data is another research direction.

\section*{Acknowledgment}
We thank the insightful comments from anonymous reviewers.
This work is supported in part by NSF under grants III-1763325, III-1909323, III-2106758, and SaTC-1930941.
Jing Ma was supported by HKBU direct grant (Ref. AIS 21-22/02).


\bibliographystyle{IEEEtran}
\bibliography{bigdata}

\end{document}